\documentclass[10pt,twocolumn,letterpaper]{article}

\usepackage{cvpr}

\usepackage{graphicx}
\usepackage{amsmath}
\usepackage{amssymb}
\usepackage{booktabs}
\usepackage{multirow}
\usepackage[pagebackref,breaklinks,colorlinks]{hyperref}

\usepackage[capitalize]{cleveref}
\crefname{section}{Sec.}{Secs.}
\Crefname{section}{Section}{Sections}
\Crefname{table}{Table}{Tables}
\crefname{table}{Tab.}{Tabs.}

\def\cvprPaperID{10999}
\def\confName{CVPR}
\def\confYear{2022}

\begin{document}

\title{Single-Stage 3D Geometry-Preserving Depth Estimation Model Training on Dataset Mixtures with Uncalibrated Stereo Data}

\author{Nikolay Patakin\textsuperscript{1,2}, Mikhail Romanov\textsuperscript{1}, Anna Vorontsova\textsuperscript{1}, Mikhail Artemyev\textsuperscript{1}, Anton Konushin\textsuperscript{1}\\
\textsuperscript{1}Samsung AI Center, Moscow \textsuperscript{2}Skolkovo Institute of Science and Technology (Skoltech)\\
{\tt\small  \{n.patakin, m.romanov, a.vorontsova, m.artemyev, a.konushin\}@samsung.com}}

\maketitle

\begin{abstract}
Nowadays, robotics, AR, and 3D modeling applications attract considerable attention to single-view depth estimation (SVDE) as it allows estimating scene geometry from a single RGB image. Recent works have demonstrated that the accuracy of an SVDE method hugely depends on the diversity and volume of the training data. However, RGB-D datasets obtained via depth capturing or 3D reconstruction are typically small, synthetic datasets are not photorealistic enough, and all these datasets lack diversity. The large-scale and diverse data can be sourced from stereo images or stereo videos from the web. Typically being uncalibrated, stereo data provides disparities up to unknown shift (\textit{geometrically incomplete} data), so stereo-trained SVDE methods cannot recover 3D geometry. It was recently shown that the distorted point clouds obtained with a stereo-trained SVDE method can be corrected with additional point cloud modules (PCM) separately trained on the geometrically complete data. On the contrary, we propose GP\textsuperscript{2}, General-Purpose and Geometry-Preserving training scheme, and show that conventional SVDE models can learn correct shifts themselves without any post-processing, benefiting from using stereo data even in the geometry-preserving setting. Through experiments on different dataset mixtures, we prove that GP\textsuperscript{2}-trained models outperform methods relying on PCM in both accuracy and speed, and report the state-of-the-art results in the general-purpose geometry-preserving SVDE. Moreover, we show that SVDE models can learn to predict geometrically correct depth even when geometrically complete data comprises the minor part of the training set.
\end{abstract}

\section{Introduction}
\label{sec:intro}

Single-view monocular depth estimation (SVDE) aims at estimating a dense depth map corresponding to an input RGB image. Being a fundamental problem in computer vision and visual understanding, it has numerous applications in autonomous driving, robotics navigation and manipulation, and augmented reality. Most applications request for approaches able to process an arbitrary RGB image and estimate a depth map that allows recovering the 3D geometry of a scene (e.g. a point cloud can be reconstructed given camera calibration).

Early SVDE methods were designed to work either in indoor scenes~\cite{nyuv2} or in autonomous driving scenarios~\cite{kitti2012}, thus being domain-specific. Accordingly, these methods demonstrate poor generalization ability in cross-domain experiments \cite{midas}. Overall, it is empirically proved that the generalization ability of SVDE models heavily depends on the diversity of the training data ~\cite{midas, LeReS, diw, wang2019web, megadepth}.

Recently, efforts have been made to obtain depth data from new sources: from computer simulation~\cite{sintel, suncg, synthia}, via 3D reconstruction~\cite{megadepth}, or from stereo data. Among them, stereo videos \cite{wang2019web, midas} and stereo images \cite{redweb, Holopix, Xian_2020_CVPR} collected from the web are the most diverse and accessible data sources. However, the depth data obtained from stereo videos is geometrically incomplete since stereo cameras' intrinsic and extrinsic parameters are typically unknown. Respectively, disparity from a stereo pair can be computed up to unknown shift and scale coefficients (UTSS). Such depth data can be used as a proxy for a ground truth depth map, but is insufficient to restore 3D geometry.

As a result, modern general-purpose SVDE methods trained on stereo data output predictions that cannot be used to recover 3D geometry \cite{midas, Xian_2020_CVPR, Lienen2021MonocularDE}. Hence, we describe these methods as \textit{not geometry-preserving} and their predictions as \textit{geometrically incorrect}. The only SVDE method that is both general-purpose and \textit{geometry-preserving} is LeReS \cite{LeReS}. It addresses geometry-preserving depth estimation with a multi-stage pipeline that recovers shift and focal length through predicted depth post-processing. This post-processing is performed using trainable point-cloud modules (PCM), thus resulting in significant overhead. Moreover, training PCM requires geometrically complete data with known camera parameters which limits the possible sources of training data.

In this work, we introduce GP\textsuperscript{2}, an end-to-end training scheme and show that SVDE models can learn correct shift themselves and benefit from training on geometrically incomplete data while being geometry-preserving.

To obtain geometrically correct depth predictions, we do not employ additional models like PCM in LeReS \cite{LeReS}. Instead, we use geometrically complete data to encourage an SVDE model to predict geometrically correct depth maps while taking advantage of diverse and large-scale UTSS data. Thus, the proposed training scheme can be applied to make any SVDE model general-purpose and geometry-preserving. To prove this, we train different SVDE models using GP\textsuperscript{2} and evaluate them on unseen datasets.

We demonstrate that the LeReS SVDE model trained with our scheme outperforms the original LeReS+PCM \cite{LeReS} trained on the same data. Then, we improve this result by switching to the LRN-based \cite{lightweightrefinenet} SVDE model, thus pushing the boundaries of general-purpose geometry-preserving SVDE even further.

Overall, our contribution is three-fold:
\begin{itemize}
    \item In this paper, we present an end-to-end training scheme GP\textsuperscript{2} that allows training a general-purpose geometry-preserving SVDE method;
    \item Through multiple zero-shot cross-dataset trials, we show that our training scheme improves both depth estimates and point cloud reconstructions obtained with the existing SVDE methods. As a result, we set a new state-of-the-art in the general-purpose geometry-preserving SVDE with an SVDE model being much faster on inference than competing methods that use PCM;
    \item In ablation studies, we show that an SVDE model learns to predict geometrically correct depth even when trained mainly on UTSS data with a small portion of the geometrically complete data.
\end{itemize}

The rest of the paper is organized as follows: in Sec.~\ref{sec:related}, we discuss the relationship between depth data and 3D geometry and give an overview of existing general-purpose SVDE approaches. Our training scheme is introduced in Sec.~\ref{sec:scheme}. Sec.~\ref{sec:eval} is dedicated to experiments: there, we describe our experimental protocol, report quantitative results, and demonstrate visualizations.

\section{Related Work}\label{sec:related}

SVDE methods have been actively investigated for decades and have significantly evolved over this period. Early depth estimation methods employed complicated heuristics to process hand-crafted features~\cite{saxena, hoiem2005}. Recently, deep learning-based approaches have been adopted for solving a wide range of computer vision tasks, including depth estimation. 

As in other computer vision problems, deep learning-based SVDE methods rely heavily on the diversity and the amount of training data. Significant work has been done to introduce new datasets, but each of them has its own biases and limitations depending on data acquisition methods. The majority of modern SVDE methods formulate depth estimation as dense pixel-wise labelling in continuous space~\cite{eigen2015predicting,laina2016deeper,cao2017estimating,nekrasov2018real}. Traditionally, depth estimation models were trained using closed-domain sensor-captured datasets with absolute depth measurements, such as NYU~\cite{nyuv2} and KITTI~\cite{kitti2012, kitti2015}. Such models overfit to the training domain and do not generalize well~\cite{midas}.

\subsection{Depth Data and 3D Geometry}

Not all applications that use depth information require absolute depth measurements. For instance, depth ordinal rankings can be sufficient to imitate the bokeh effect for image processing or to estimate occlusion boundaries in AR applications. Ordinal SVDE methods address depth estimation by comparing the depth of each two pixels. Accordingly, the training data might contain only ordinal rankings for several pixels of each depth map; this data can be manually labeled by showing random pixel pairs to an assessor and asking which pixel in a pair is closer~\cite{diw}.

Single-view depth estimation can also be reformulated as predicting inverse depth up to unknown scale and shift (UTSS)~\cite{midas}. UTSS data can be obtained from a stereo image through stereo matching even when calibration camera parameters are unknown. Hence, learning to estimate depth up to unknown shift and scale opens opportunities for training SVDE models on large-scale and diverse stereo data collected from the web~\cite{midas, wang2019web,redweb}. UTSS predictions are geometrically incomplete and insufficient to restore 3D geometry, yet they can be used in the same scenarios as depth ordinal rankings.

Recovering 3D geometry is necessary for such applications as one-shot 3D photography~\cite{kopf3d}. Accordingly, these applications require depth estimates that can be either absolute or known up-to-scale (UTS). In the UTS setting, we assume the entire scene to be uniformly scaled by an unknown factor, so estimated UTS depth preserves 3D geometry and keeps the objects shapes unchanged. Ground truth UTS depth maps can be obtained by a combination of structure-from-motion (SfM) and multi-view stereo (MVS) methods given a multi-view image collection~\cite{megadepth}. However, this scheme allows processing only static objects: the ground truth depth cannot be obtained for moving objects such as people, cars, or animals.

\subsection{General-purpose SVDE}\label{sec:GP-SVDE}

\textbf{MiDaS.} In MiDaS~\cite{midas}, multiple geometrically complete (UTS) and geometrically incomplete (UTSS) datasets are mixed together to obtain diverse and large-scale training data. Scale-and-shift invariant loss functions were used to address stereo datasets geometric incompleteness. As a result, the MiDaS model outputs UTSS predictions, so 3D geometry cannot be recovered. 

\begin{figure*}[ht!]
    \centering
    \includegraphics[width=0.95\textwidth]{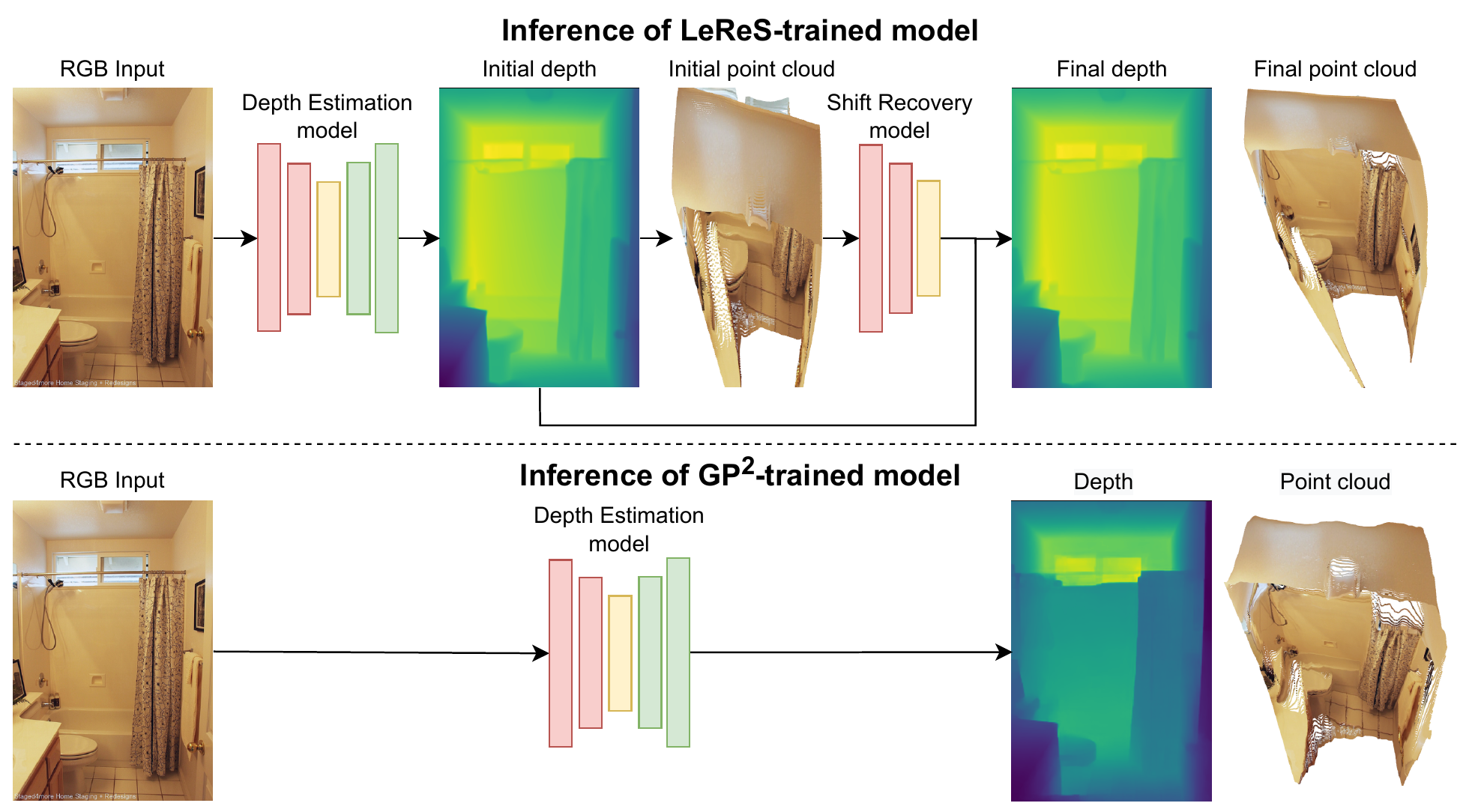}
    \caption{Our approach compared to the only existing general-purpose geometry-preserving depth estimation method. Being trained with our GP\textsuperscript{2} scheme, conventional SVDE models can learn correct shift values themselves and do not require any additional post-processing.}
    \label{fig:teaser}
\end{figure*}

\textbf{LeReS.} To the best of our knowledge, the only existing general-purpose and geometry-preserving SVDE method is LeReS~\cite{LeReS}. It estimates geometrically correct depth maps through a multi-stage pipeline (see Fig. \ref{fig:teaser}). This pipeline includes an SVDE model and two point cloud modules (PCM). The SVDE model is trained on a mixture of absolute and UTSS data and outputs initial UTSS depth estimates that are refined with PCM. Specifically, one point cloud module corrects depth shift, while another estimates camera focal length. Both modules are essential for estimating geometrically correct depth.

It is worth noting, that the LeReS SVDE model is trained to predict UTSS depth, while stereo data provides UTSS disparity. UTSS disparity cannot be straightforwardly converted to UTSS depth, as adding a constant to a disparity map corresponds to a non-linear depth map transformation. Accordingly, LeReS uses ranking loss functions on uncalibrated stereo data. This leads to an inefficient use of the training data, since metric information is then missed: two ranked points might be either almost equidistant or be far from each other.

UTSS losses computed in disparity space allow to utilize stereo data more efficiently~\cite{midas}. In this case, UTSS disparities are to be predicted. To reconstruct a point cloud from such an estimate, disparity shift needs to be recovered. Since the LeReS SVDE model outputs UTSS depth, PCM yields a single additive constant in depth domain. Accordingly, it cannot be used to perform the required non-linear depth correction, thus being inapplicable for correcting predictions of stereo-trained SVDE models~\cite{midas, DPT}. 

Besides aforementioned drawbacks, LeReS requires calibrated data for training PCM. In contrast, our training scheme allows making geometrically correct estimates even if camera parameters are unknown. In other words, GP\textsuperscript{2} imposes weaker restrictions on the training data that opens up opportunities for training on large-scale and diverse data. Furthermore, while LeReS addresses geometry-preserving SVDE with a multi-stage pipeline, we solve the same problem with a single-stage approach.

\section{Proposed Training Scheme}\label{sec:scheme}

The key observation behind our scheme is that SVDE models that use UTSS data for training can still learn to make geometrically correct predictions out of UTS samples present in the training mixture. Existing general-purpose methods trained with UTSS data use only UTSS~\cite{midas, DPT, LeReS} or ranking losses~\cite{LeReS}, thus missing helpful information about correct depth shifts contained in geometrically complete data. On the contrary, we use this information during training to enforce the geometrical correctness of depth estimates. 

\textbf{Mixing datasets.} To train an SVDE model on the mixture of UTS and UTSS data, we propose using a combination of a scale-invariant (UTS) loss and a shift-and-scale-invariant (UTSS) loss; we use the former whereas possible. More formally, we train SVDE models using the following loss function:
\begin{equation}
     \mathcal{L}_{Mixture} = \mathbb{I}_{UTS} \mathcal{L}_{UTS} + \mathcal{L}_{UTSS},
     \label{eq:mixture}
\end{equation}
where $\mathbb{I}_{UTS}=1$ for UTS samples and $0$ for UTSS samples. The UTSS loss forces the SVDE model to generalize to the diverse UTSS stereo datasets, while the UTS loss encourages producing geometrically correct estimates.

\textbf{Choosing predictions space.} Uncalibrated stereo data provides inverse-depth, or \textit{disparity}, up to unknown shift and scale; therefore, it cannot be converted into depth straightforwardly. Thus, UTSS SVDE models are typically trained to predict disparities~\cite{midas, DPT}. On the contrary, we aim at training an up-to-scale SVDE model, so we follow best practices for UTS models while choosing predictions space. The advantages of training UTS models to predict depth in logarithmic space are proved empirically~\cite{megadepth, dorn, mannequin, kopf3d}, and a number of loss functions for log-depth space are proposed~\cite{eigen, megadepth}. Eventually, we opt for making log-depth predictions.

\textbf{Shift- and scale-invariant loss.} The existing shift- and scale-invariant loss functions~\cite{wang2019web, midas, LeReS} are based on $L_1$ loss function with additional quantile trimmings or $tanh$-based adjustments~\cite{LeReS}. In this study, we want to mitigate the effects of using elaborate loss functions, so we validate our training scheme in the most straightforward and demonstrative experimental setting. Specifically, we use a pure $L_1$ loss function without any additional terms.

Let $l$ denote log-depth estimated by the model, $d~=~\exp(l)$ --  estimated depth, and $d^*$ -- ground truth depth (for geometrically complete datasets). We can express disparity as $D = \frac{1}{d} = exp(-l)$, and ground truth disparity as $D^* = \frac{1}{d^*}$. Before computing loss function, we align the mean and standard deviation of predictions and ground truth values to make it shift- and scale-invariant:
\begin{equation}
    \hat{D} = \frac{D - \mu_D}{\sigma_D}  \quad \quad \quad \hat{D}^* = \frac{D^* - \mu_{D^*}}{\sigma_{D^*}}    
\end{equation}

Then, the UTSS loss function for a single sample can be formulated as:
\begin{equation}
    \mathcal{L}_{UTSS} = \frac{1}{N} \sum \limits_{i=1}^N  \left| \hat{D}_i - \hat{D}^*_i \right|,
    \label{eq:loss_uts}
\end{equation}

\textbf{Scale-invariant UTS loss.} Such basic functions as $L_1$ or $L_2$, can be used as scale-invariant UTS loss functions. But, a number of more elaborate task-specific scale-invariant loss functions have been introduced: specifically, Multi-Scale Gradient Loss \cite{megadepth}, and Pairwise Loss \cite{eigen}. 

In our study, we calculate scale-invariant loss function in log-depth space, beforehand we align median of log-differences:
\begin{equation}
    \mu = median(l - \log d^*)
\end{equation}

and UTS loss function can be written as:
\begin{equation}
    \mathcal{L}_{UTS} = \frac{1}{N} \sum \limits_{i=1}^N \left| l_i - \log d^*_i - \mu \right|,
    \label{eq:l1_uts}
\end{equation}

Our training scheme does not depend on the choice of a loss function, so other scale-invariant and shift- and scale-invariant losses can be used instead of scale-invariant and shift- and scale-invariant pointwise $L_1$.

\section{Experiments}\label{sec:eval}

This section is dedicated to experiments: we describe our experimental protocol, report quantitative results, and present visualizations.
We show the effectiveness of the proposed GP\textsuperscript{2} scheme against multi-stage LeReS approach, as well as against methods that use only UTS data for training.
In ablation studies, we show that even when UTS data comprises a minor part of a training set, it is sufficient to train the geometry-preserving SVDE method.

\subsection{SVDE Models}

First, we conduct experiments with the SVDE model from LeReS \cite{LeReS}.
To validate the GP\textsuperscript{2} scheme with different architectures, we train models based on the Light-weight Refine Net \cite{lightweightrefinenet} meta-architecture. For an efficient light-weight model, we use EfficientNet-Lite0~\cite{efficientnet-lite} as a backbone, and for a more powerful model, we choose EfficientNet-B5 ~\cite{efficientnet}. Hereinafter, we refer to these LRN-based SVDE models as to Lite0-LRN and B5-LRN, respectively.

\begin{table*}[!htbp]
    \centering \setlength{\tabcolsep}{2.5pt}
    \begin{tabular}{l|c|c|ccccc|c}
    \multirow[l]{2}{*}{Training dataset mixture} & \multirow[l]{2}{*}{Method} & \multirow[l]{2}{*}{Training scheme} & TUM & NYU & Sintel & ETH3D & iBims-1 & \multirow[l]{2}{*}{Average rank$\downarrow$} \\
	& & & $\delta_{1.25}\downarrow$ & $\delta_{1.25}\downarrow$ & $\text{rel}\downarrow$ & $\text{rel}\downarrow$ & $\text{rel}\downarrow$ & \\
	\hline
	MD & \multicolumn{2}{c|}{MegaDepth~\cite{megadepth}} & 0.331& 0.344	& 0.490 & 0.276	& 0.389 &  12\\
    MannequinChallenge & \multicolumn{2}{c|}{Li et al.~\cite{mannequin}} & 0.226 & 0.239 & 0.431 & 0.249 & 0.175 & 10.8 \\
    MIX6~\cite{DPT} $\to$ NYU~\cite{nyuv2} & \multicolumn{2}{c|}{DPT~\cite{DPT}} & 0.151 & \textbf{\textit{0.049}}* & 0.389 & 0.189 & 0.101 & 5.2\\
    \hline
    \multirow[l]{3}{*}{DIML\textsubscript{I}+MD (only UTS data)} 
    & Lite0-LRN & UTS-only & 0.170 & 0.142 & 0.428 & 0.191 & 0.117 & 9\\
    & B5-LRN & UTS-only & 0.145 & \underline{0.106} & 0.397 & 0.174 & 0.100 & 4.4\\
    & LeReS & UTS-only & 0.146 & 0.123 & 0.417 & 0.174 & 0.108 & 6.4\\
    \hline
    \multirow[l]{3}{*}{DIML\textsubscript{I}+RW+MD+MV} & Lite0-LRN & GP\textsuperscript{2} & 0.144 & 0.142 & 0.354 & 0.177 & 0.115 & 6.4\\
    & B5-LRN & GP\textsuperscript{2} & \textbf{0.130} & \underline{0.106} & \underline{0.328} & 0.154 & \underline{0.098} & \textbf{2.2} \\	
    & LeReS & GP\textsuperscript{2} & \underline{0.139} & 0.117 & 0.363 & 0.175 & 0.110 & 5.4\\
	\hline
    \multirow[l]{3}{*}{LeReS mixture} & LeReS+PCM~\cite{LeReS} & LeReS & 0.193 & 0.139 & 0.403 & 0.172 & 0.140 & 7.8\\
    & B5-LRN & GP\textsuperscript{2} & 0.237 & \textbf{0.095} & \textbf{0.320} & \textbf{0.131} & \textbf{0.092} & \underline{3.2} \\
    & LeReS & GP\textsuperscript{2} & 0.192 & 0.114 & 0.343 & \underline{0.146} & 0.103 & 4.6
    \end{tabular}
    \caption{Results of SVDE models trained on different dataset mixtures with and without GP\textsuperscript{2}. The state-of-the-art results for are marked bold, the second best are underlined.\\
    \textsuperscript{*} This model was fine-tuned on the training subset of NYUv2. Other models had never seen any NYUv2 samples during training.}
    \label{tab:results}
\end{table*}

\subsection{Dataset Mixtures} To demonstrate versatility of our training scheme, we train SVDE models on two datasets mixtures introduced in MiDaS \cite{midas} and LeReS \cite{LeReS}, respectively. 

Not all the datasets presenting in these mixtures can be used 'as is' due to reproducibility issues. Some of these datasets contain data in a form of web links referring to the sources which are not publicly available~\cite{midas}, or has been deleted ever since~\cite{wang2019web}; while some other datasets do not contain depth data explicitly, so disparities should be extracted through some unknown stereo-matching protocol~\cite{diml, Holopix, midas}. In general, we aim at reproducing dataset mixture as close to the original ones as possible; however, they are non-equivalent.

\textbf{MiDaS training data.} Similar to MiDaS, we train SVDE models on a mixture of four datasets: MegaDepth~\cite{megadepth} (denoted as MD for brevity; contains UTS data), DIML Indoor~\cite{diml} (DIML\textsubscript{I}, UTS), RedWeb~\cite{redweb} (RW, UTSS), and Stereo Movies (MV, UTSS). We do not use WSVD dataset~\cite{wang2019web} from the original mixture. Differently, we use 49 stereo movies (listed in Supplementary) against 23 used in the original MiDaS. Overall, our Stereo Movies dataset contains $\approx$140k samples against 75k samples in \cite{midas}. Hereinafter, we refer to the obtained dataset mixture as to DIML\textsubscript{I}+RW+MD+MV.

\textbf{LeReS training data.} The LeReS dataset mixture contains Taskonomy ~\cite{Taskonomy} (absolute depths), 3D Ken Burns~\cite{KenBurns} (absolute), DIML Outdoor~\cite{diml} (absolute), HoloPix~\cite{Holopix} (UTSS), and HRWSI~\cite{Xian_2020_CVPR} (UTSS). Below, we denote this dataset mixture as the LeReS mixture. 

We evaluate all SVDE models on several datasets unseen during training: the NYU test set, the split of TUM RGB-D proposed by Li et al.~\cite{mannequin}, the synthetic Sintel dataset, ETH3D, and iBims-1. For the ETH3D dataset, we follow MiDaS, rendering ground truth depth maps from reconstructed point clouds.

\begin{figure*}[ht!]
    \centering
\setlength{\tabcolsep}{1pt}\begin{tabular}{cccccc} 
Image                                            & MegaDepth \cite{megadepth}                       & Li et al. \cite{mannequin}                       & MiDaSv2 \cite{midas}                             & LeReS+PCM\cite{LeReS}                            & Ours, B5-LRN                                     \\
\includegraphics[width=0.1625\linewidth]{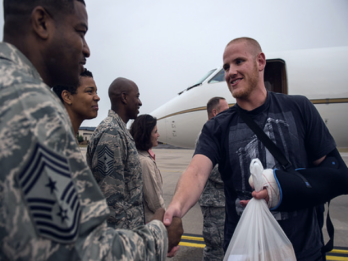} &
\includegraphics[width=0.1625\linewidth]{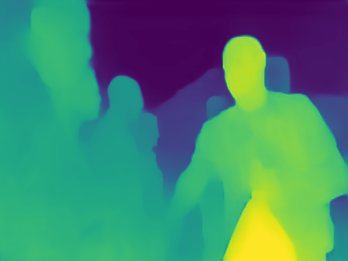} & 
\includegraphics[width=0.1625\linewidth]{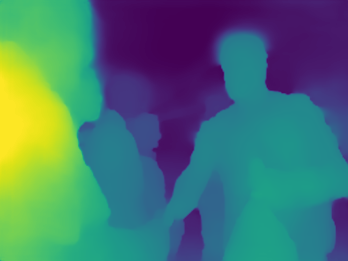} &
\includegraphics[width=0.1625\linewidth]{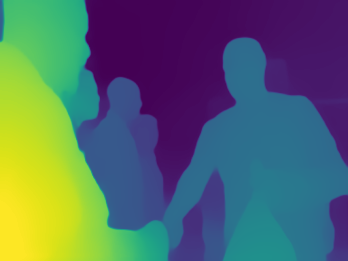} & 
\includegraphics[width=0.1625\linewidth]{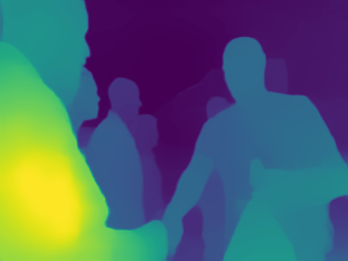} &
\includegraphics[width=0.1625\linewidth]{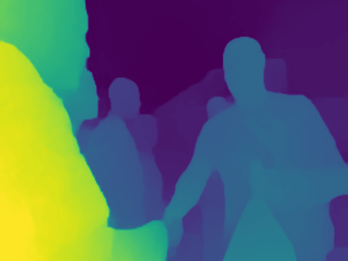} \\
\includegraphics[width=0.1625\linewidth]{images/fig_depth/image-006.png} &
\includegraphics[width=0.1625\linewidth]{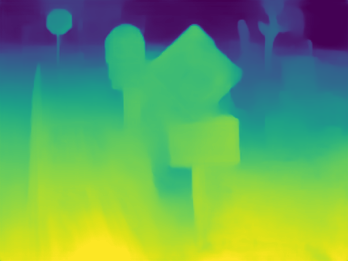} & 
\includegraphics[width=0.1625\linewidth]{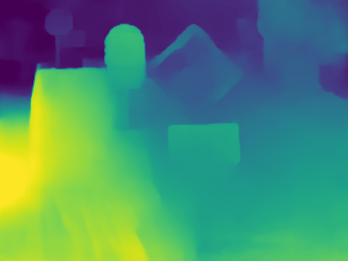} &
\includegraphics[width=0.1625\linewidth]{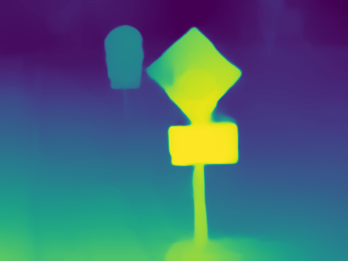} & 
\includegraphics[width=0.1625\linewidth]{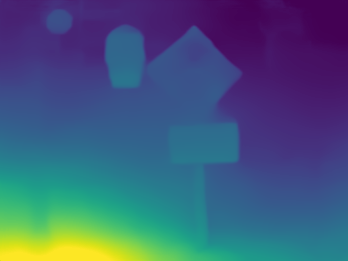} &
\includegraphics[width=0.1625\linewidth]{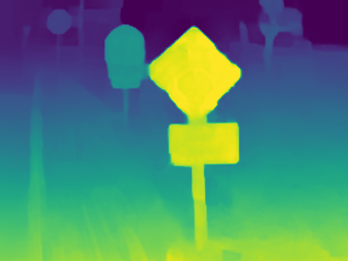} \\
\includegraphics[width=0.1625\linewidth]{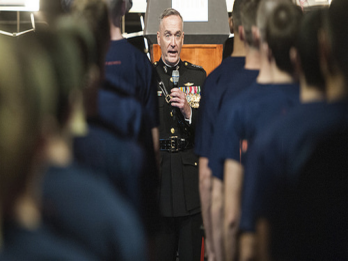} &
\includegraphics[width=0.1625\linewidth]{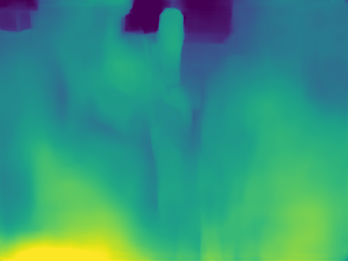} & 
\includegraphics[width=0.1625\linewidth]{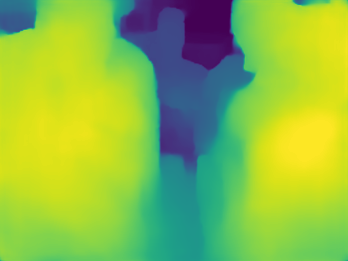} &
\includegraphics[width=0.1625\linewidth]{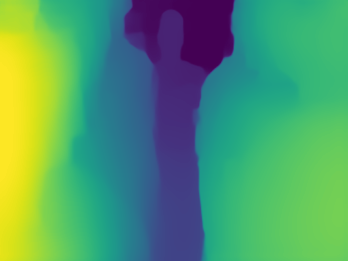} & 
\includegraphics[width=0.1625\linewidth]{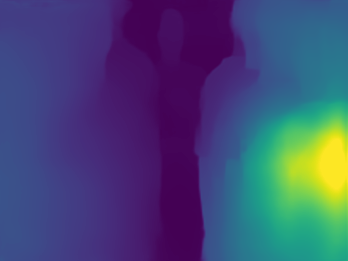} &
\includegraphics[width=0.1625\linewidth]{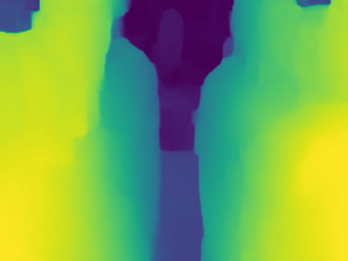}
\end{tabular}
    \caption{Depth maps estimated with existing SVDE methods and our GP\textsuperscript{2}-trained B5-LRN model.}
    \label{fig:visualization}
\end{figure*}

\begin{table*}[ht!]
    \centering \setlength{\tabcolsep}{2.5pt}
    \begin{tabular}{l|c|c|cc|c}
	Training dataset mixture & Method & Training scheme & $\delta_{1.25}\downarrow$ & $\text{rel}\downarrow$ & Point cloud RMSE $\downarrow$ \\
	\hline
    \multirow[l]{3}{*}{LeReS mixture} & B5-LRN+PCM & GP\textsuperscript{2} & \textbf{0.127} & \textbf{0.113} & \textbf{0.363} \\
	& LeReS+PCM & GP\textsuperscript{2} & 0.161 & 0.137 & 0.395 \\
	& LeReS+PCM & LeReS & 0.216 & 0.164 & 0.429 	
\end{tabular}
    \caption{Results of depth estimation and point cloud reconstruction on the 2D-3D-S dataset for methods trained with and without GP\textsuperscript{2}. The GP\textsuperscript{2}-trained methods use pretrained PCM from LeReS only for camera focal length estimation.}
    \label{tab:results_pc}
\end{table*}

\subsection{Experimental Protocol}

\textbf{Implementation details.} We train LRN-based models with Radam~\cite{radam} optimizer with LookAhead~\cite{lookahead}. The learning rate is $0.001$, and each batch consists of 24 samples. We select samples from each training dataset with equal probability, as proposed in MiDaS~\cite{midas}, which helps to avoid domain biases caused by unequal datasets sizes. The training takes 40 epochs with $10{,}000$ training steps per epoch. During training, we use the same set of augmentations: resizing, padding, taking random crops to obtain samples of size 384$\times$384, geometrical transformations, and color distortions. During inference, we resize input images so their smaller side becomes $384$ pixels. 

We reproduce the LeReS training procedure using the official code for a fair comparison. Specifically, we train the SVDE model using SGD with a batch size of 40, an initial learning rate of 0.02, and a learning rate decay of 0.1. Input images are resized to 448$\times$448, and flipped horizontally with a probability of $0.5$. 

All methods are implemented with Python and PyTorch~\cite{pytorch}. For training and evaluation, we used Nvidia Tesla P40, while performance tests were run on a Nvidia GeForce GTX 1080 Ti. 

\textbf{Point cloud reconstruction.} To obtain a point cloud from a depth map, camera parameters should be known. We assume that the RGB camera used to capture data can be described in terms of a pinhole model. Accordingly, the unprojection from 2D coordinates and depth to 3D points can be performed as:
\begin{equation}
\begin{split}
    x = \frac{u - u_0}{f} d, \
    y = \frac{v - v_0}{f} d, \
    z = d
\end{split}
\end{equation}
where $(u_0, v_0)$ are the camera optical center, $f$ is the focal length, and $d$ is the depth. The focal length affects the point cloud shape as it scales $x$ and $y$ coordinates, but not $z$. We assume that the camera optical center is the geometric center of an image, so only focal distance should be estimated. Following original LeReS, we estimate it via PCM.

\textbf{Metrics.} We follow the MiDaS evaluation protocol, using $\delta_{1.25}$ for NYU and TUM RGB-D and $\mathrm{rel}$ for Sintel, ETH3D, and iBims-1. We measure the quality of point cloud reconstruction with pointwise RMSE instead of LSIV proposed in LeReS. We argue that it is a more transparent and straightforward approach since it does not require any additional modalities such as segmentation maps in LSIV.

\subsection{Quantitative Comparison}

In this section, we demonstrate the efficiency of our training scheme for depth estimation and point cloud reconstruction.

\textbf{Depth estimation.} We present the results of our SVDE models trained on DIML\textsubscript{I}+RW+MD+MV and LeReS mixtures in Tab. \ref{tab:results}.

According to average ranks, the GP\textsuperscript{2}-trained LeReS SVDE model outperforms the original LeReS+PCM while using the same training data. Moreover, GP\textsuperscript{2} allows making UTS predictions with a simple single-stage pipeline without any post-processing such as PCM.

As one might see, training on the UTSS part of DIML\textsubscript{I}+RW+MD+MV mixture via GP\textsuperscript{2} is profitable for all examined SVDE models as compared to the baselines trained on the UTS data only.

Also, we claim our GP\textsuperscript{2}-trained B5-LRN to set a new state-of-the-art in the general-purpose geometry-preserving SVDE. Being trained with GP\textsuperscript{2} on the LeReS mixture, B5-LRN is the most accurate on NYU, Sintel, ETH3D, and iBims-1. At the same time, our B5-LRN model trained on the DIML\textsubscript{I}+RW+MD+MV mixture via GP\textsuperscript{2} has the best average rank among all models.

Note that since our SVDE models are trained to estimate UTS depth, we compare them to other UTS SVDE methods. Such well-known stereo-trained methods as DPT~\cite{DPT} (trained on MIX6~\cite{DPT} only) and MiDaS~\cite{midas} predict UTSS disparity, so the shift of these estimates cannot be corrected via shift recovery, as discussed in Sec.~\ref{sec:GP-SVDE}. Accordingly, we do not perform quantitative evaluation against SVDE methods predicting UTSS disparity, yet we present the qualitative comparison in Fig.~\ref{fig:visualization} and supplementary. Authors of DPT~\cite{DPT} released several models, including the one fine-tuned on the training subset of NYUv2 to estimate absolute depth. In Tab.~\ref{tab:results}, we report the results of the fine-tuned DPT model.

There, we visualize depth estimates obtained by several existing SVDE methods (Li et. al~\cite{mannequin}, MegaDepth~\cite{megadepth}, MiDaS~\cite{midas}, LeReS+PCM~\cite{LeReS}) and our GP\textsuperscript{2}-trained B5-LRN model. 

\begin{figure*}[ht!]
    \centering
        \includegraphics[width=1\linewidth]{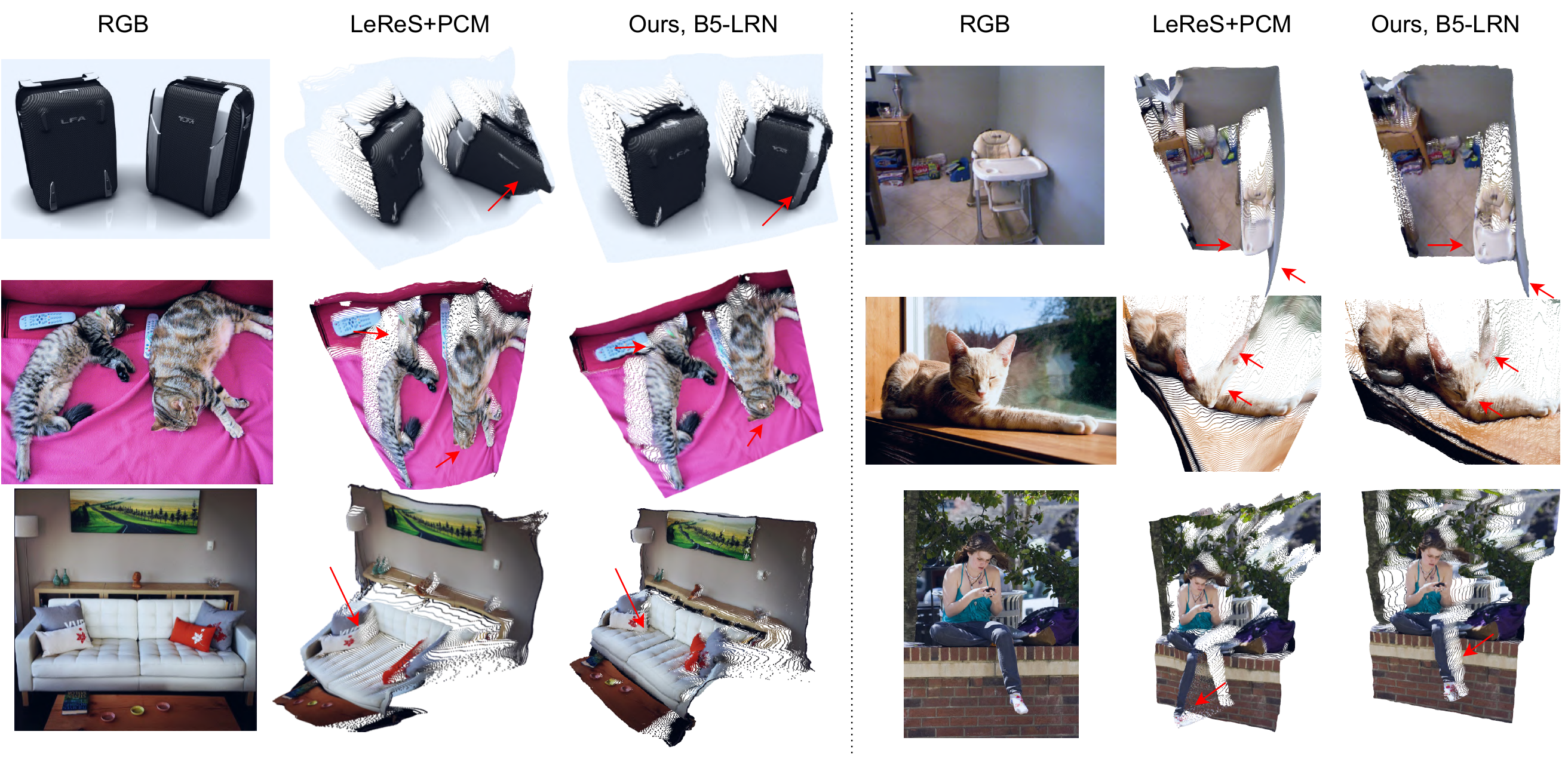}
    \caption{Point clouds reconstructed from depth maps estimated by the original LeReS+PCM and our GP\textsuperscript{2}-trained B5-LRN using PCM for focal recovery. Red arrows are pointing on areas which appear to be difficult for LeReS+PCM, while our method recovers 3D geometry nicely in these areas.}
    \label{fig:pointclouds}
\end{figure*}

\textbf{Point cloud reconstruction}. It is worth noticing that geometry-preserving SVDE implies predicting depth maps that allow recovering 3D geometry, but it does not imply recovering 3D geometry itself. Accordingly, we do not propose any point cloud reconstruction techniques and use the LeReS PCM for focal length estimation instead.  Importantly, our SVDE pipeline does not require PCM. We reconstruct point clouds only for a comprehensive comparison with LeReS+PCM. 

In Tab.~\ref{tab:results_pc}, we report point cloud accuracy on the 2D-3D-S dataset~\cite{stanford3d} unseen during training. Since both LeReS and our GP\textsuperscript{2}-trained SVDE model rely on the same PCM, the observed RMSE improvement should be attributed to the more accurate depth estimation.

The examples of reconstructed point clouds are visualized in Fig. \ref{fig:pointclouds}. We identify the areas where the original LeReS+PCM outputs erroneous predictions while our GP\textsuperscript{2}-trained B5-LRN estimates 3D geometry adequately (depicted with red arrows).

\subsection{Ablation Study}

\textbf{UTS / UTSS ratio}. In ablation studies, we show that even when UTS data comprises a minor part of a training set, it is sufficient to train the geometry-preserving SVDE method. To prove that, we conduct experiments with our B5-LRN model on the NYU Raw dataset \cite{nyuv2}. Since it contains absolute depth maps, we can make them either UTS or UTSS. To convert absolute depth to UTS, we multiply it by a random positive coefficient. To obtain UTSS data, we multiply the inverse depth by a random scale and then shift it by a random additive value. In this study, we divide the data into two parts and make them UTS and UTSS, respectively. To obtain comprehensive results, we run experiments with different ratios of UTS data. We compare the model trained with GP\textsuperscript{2} on the UTS and UTSS data against the same model trained only on the UTS part of data.

\begin{figure}[!b]
    \centering
        \includegraphics[height=6cm]{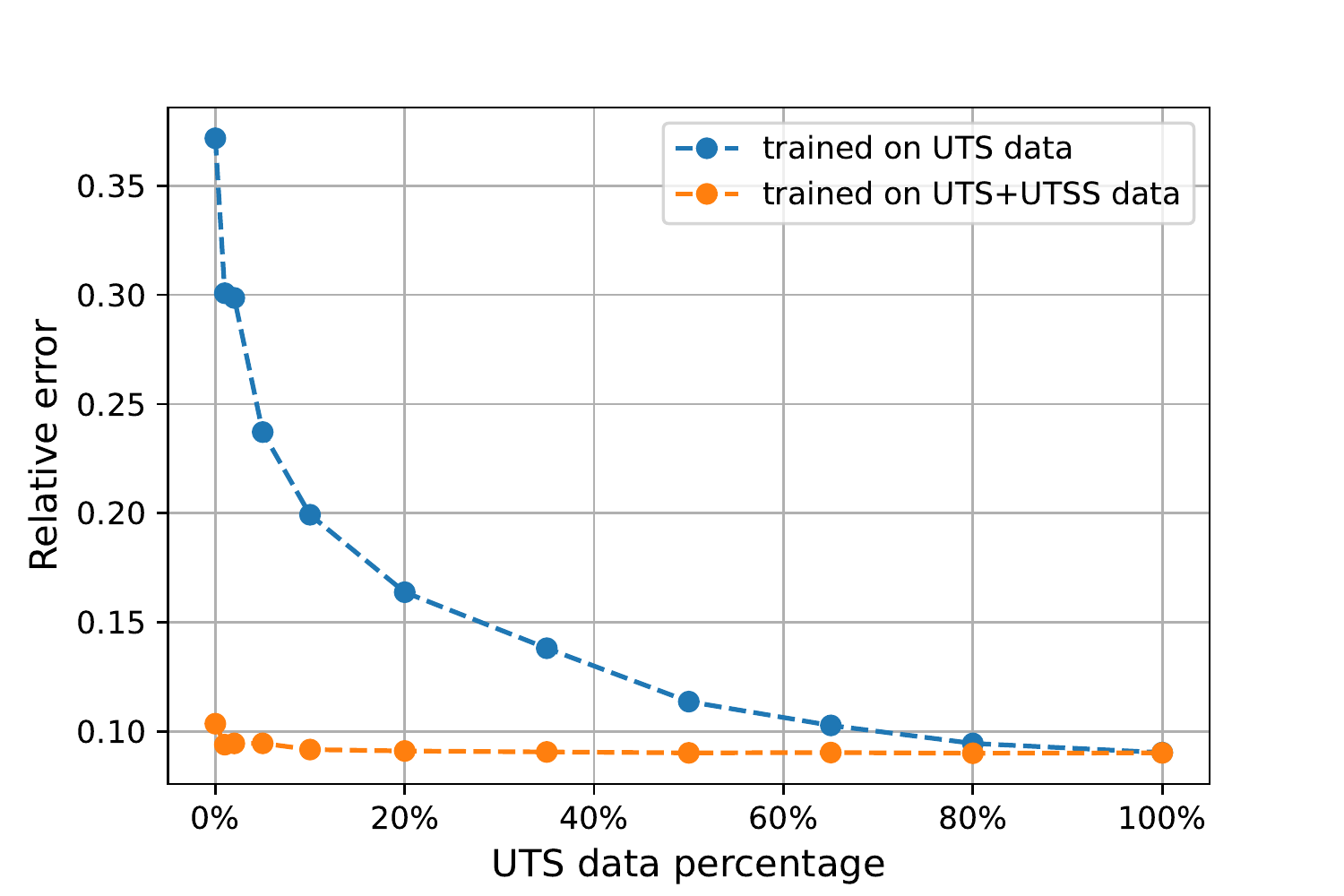}
    \caption{Results of the ablation study on the ratio of UTS and UTSS data from NYU Raw.}
    \label{fig:poc}
\end{figure}

According to Fig.~\ref{fig:poc}, the more UTS training data we use, the more accurate are the depth estimates. At the same time, the method trained on a mixture of UTS and UTSS data shows similar results for all UTS ratios; in other words, it performs as if it was trained on the dataset consisting of UTS data only. In practice, that means that the full geometrically correct dataset is not required since the proposed GP\textsuperscript{2} allows training as accurate SVDE model with only 10-20\% of UTS data.

\begin{table*}[ht!]
    \centering \setlength{\tabcolsep}{2.5pt}
    \begin{tabular}{l|c|c|c}
	Model & Backbone & Inference time, ms & Average rank$\downarrow$ \\
	\hline
    LeReS+PCM \cite{LeReS} & ResNeXt101 \cite{resnext} & 201.9 & 7.8 \\
    LeReS SVDE model & ResNeXt101 \cite{resnext} & 45.1 & 4.6 \\
    B5-LRN (ours) & EfficientNet-B5\cite{efficientnet} & 35.4 & 2.2 \\
    Lite0-LRN (ours) & EfficientNet-lite0 \cite{efficientnet-lite} & 11.1 & 6.4 \\
\end{tabular}
    \caption{Average GPU latencies for different SVDE methods.}
    \label{tab:latency_bench}
\end{table*}

\subsection{Performance}

Tab.~\ref{tab:latency_bench} shows inference time for different SVDE models with and without PCM. The reported timings were measured on a single Nvidia GeForce GTX 1080 Ti GPU with a batch size of 1. We infer RGB images in $384\times 384$ resolution and report the latency averaged over 100 runs. To prove that we do not achieve better performance at the cost of depth estimation accuracy, we also report the average dataset rank for each model (same as in Tab.~\ref{tab:results}).

Here, we do not consider the CPU time required to project an estimated depth map into a point cloud, which is necessary to run PCM. Nevertheless, the use of PCM results in a significant overhead even if only model GPU inference time is taken into account. 

Hence, if computational resources are limited, fast and efficient SVDE models that do not rely on PCM should be preferred. Accordingly, we can conclude that GP\textsuperscript{2}-trained SVDE models outperform LeReS+PCM in both depth estimation accuracy and inference speed.

\section{Limitations}

Through extensive evaluation, we demonstrate that GP\textsuperscript{2}-trained SVDE models perform well on both indoor and outdoor datasets. However, these datasets do not cover all possible use cases, so we cannot state that our method will be as accurate if given arbitrary inputs.

\section{Broader Impact}

In this paper, we aim to build a method that would push the boundaries of general-purpose geometry-preserving SVDE and become a new state-of-the-art. Hence, future research in SVDE might benefit from this paper. Our method may be potentially used in self-driving perception and navigation systems, thus improving the safety of autonomous vehicles. At the same time, there is a concern on privacy issues of the widespread deployment of modern computer vision systems. Like many computer vision algorithms that can be used for surveillance purposes, our method may harm privacy if exploited by malicious actors. Nevertheless, we believe that the positive effect on safety is more significant than a hypothetical negative effect on privacy.

\section{Conclusion}

In this paper, we presented GP\textsuperscript{2}, a novel scheme of training an arbitrary SVDE model so as it becomes general-purpose and geometry-preserving. Experiments with different SVDE models and multiple dataset mixtures proved that our training scheme improves both depth estimates and point cloud reconstructions obtained with these models. Furthermore, with an SVDE model trained with GP\textsuperscript{2}, we set a new state-of-the-art in the general-purpose geometry-preserving SVDE. Moreover, we showed that a small amount of UTS data in the mixture is sufficient to train a geometry-preserving SVDE model; this opens new opportunities for using large-scale and diverse yet uncalibrated stereo data for geometry-preserving SVDE. Through performance tests, we demonstrated that GP\textsuperscript{2}-trained SVDE models outperform the competitors not only in depth estimation accuracy but also in inference speed.

{\small
\bibliographystyle{ieee_fullname}
\bibliography{main.bib}
}

\appendix
\section{Importance of Accurate Shift and Scale for 3D Geometry Recovering}
\label{apdx:utss_depth}

In this section, we show that accurate shift is crucial for recovering 3D geometry while knowing scale value is not necessary.

For data obtained from stereo videos \cite{midas}, or stereo images \cite{redweb}, ground-truth disparity can be extracted only up to unknown scale and shift coefficients. Without knowing the correct disparity shift value, 3D geometry can not be recovered. 

Hereinafter, we assume that an RGB camera can be described in terms of a pinhole model. 

Let us consider the 3D line $l$ parametrized with coefficients $a, b, c$. For projection point $x, y$ on the camera plane, its depth can be calculated as:
\begin{equation}
    d = ax + by + c,
\end{equation}
So, we can assign a depth value for each point along this 3D line. Suppose that inverse depth (disparity) values for the points along the line are known up to shift and scale:
\begin{equation}
    \frac{1}{\tilde{d}} = \frac{C_1}{ax + by + c} + C_2,
\end{equation}
or, equivalently,
\begin{equation}
    \tilde{d} = \frac{ax + by + c}{C_1 + C_2 (ax + by + c)}.
\end{equation}

The expression above defines a 3D line if and only if $C_2 = 0$. Therefore, to obtain predictions from which 3D geometry can be recovered, a neural network should explicitly estimate the $C_2$ coefficient.

As shown, $C_2$ coefficient has a large impact on the 3D geometry. At the same time, $C_1$ affects only the scale of the scene. To illustrate that, we can consider mapping from a camera plane point $(x,y)$ having a depth $d$ to a 3D point:

\begin{equation}
\begin{pmatrix}x  \\y \\ d \end{pmatrix} \mapsto \begin{pmatrix} \frac{(x-c_x)d}{f_x} \\ \frac{(y-c_y) d}{f_y} \\ d\end{pmatrix}.
\label{eq:pinhole}
\end{equation}

Suppose that the original depth map gets scaled by a factor $C_1$. According to \ref{eq:pinhole}, the coordinates of 3D points get then multiplied by $C_1$ as well. We can interpret this as the entire scene getting scaled by $C_1$ without affecting the geometry correctness (e.g., all angles and curvatures remain unchanged). 

\section{LRN-based Neural Network}
\label{apdx:architecture}

\begin{figure}[!ht]
    \centering
    \includegraphics[width=0.45\textwidth]{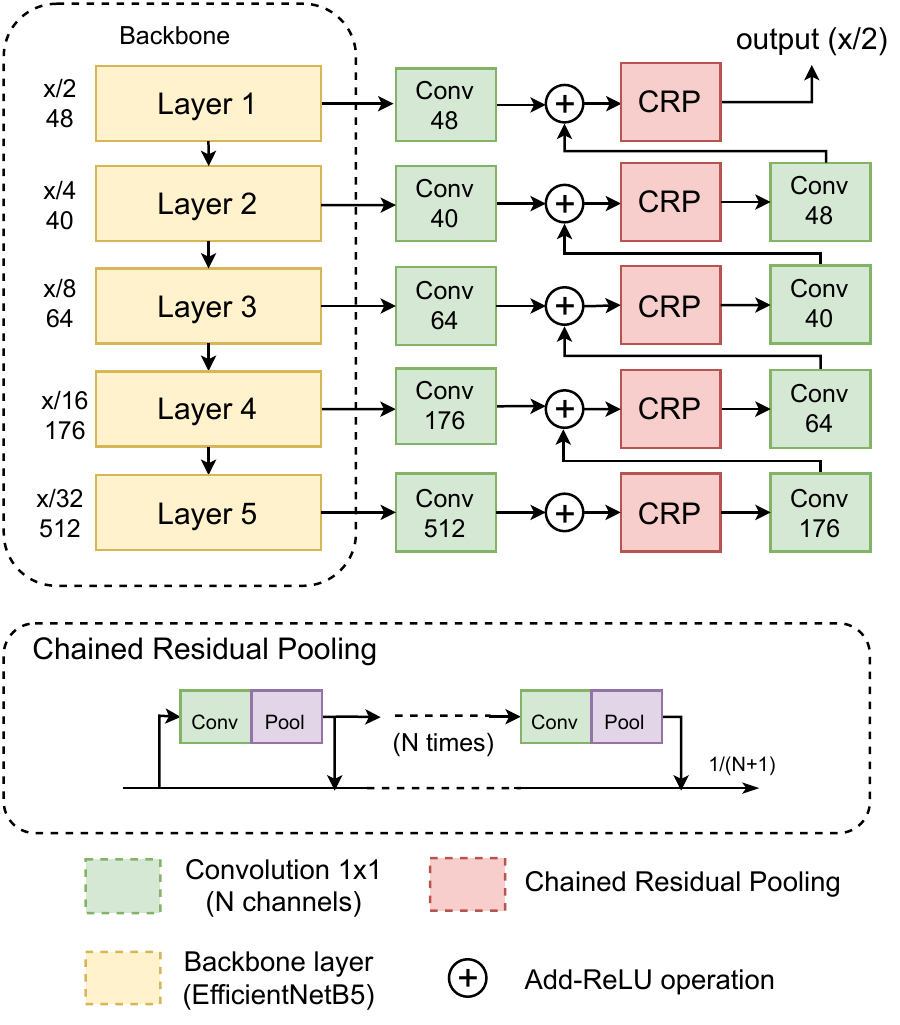}
    \caption{Architecture of our LRN-based SVDE models. }
    \label{fig:lrn}
\end{figure}

Fig. \ref{fig:lrn} depicts the architecture of our LRN-based SVDE models. Following~\cite{midas}, we use a RefineNet architecture to address the depth estimation problem. For the sake of efficiency, we use Light-Weight RefineNet (LRN)~\cite{lightweightrefinenet}.

The encoders are based on architectures from the EfficientNet family~\cite{efficientnet}: EfficientNet-Lite0 and EfficientNet-B5, both pre-trained on \emph{ImageNet}.

In the original LRN model, each encoder output is processed with a $1\times1$ convolution with $256$ output channels. This parameter is hard-coded for both light-weight and powerful models, so we can neither choose smaller values if building a light-weight decoder or use more channels for more powerful decoder architectures. We claim this non-adaptive approach to be suboptimal and propose a more flexible alternative. Unlike the original LRN, we set the number of output channels in the fusion convolutional layer equal to the number of channels in the corresponding backbone level. Then, each encoder output gets fused with the features from a deeper layer (see Fig.~\ref{fig:lrn}), and the fused output has the same number of channels as the encoder output. 

\begin{table}[h!]
    \centering
\begin{tabular}{|l|r|}
\hline
Name                                        & Frames \\
\hline
\hline
3-D Sex and Zen: Extreme Ecstasy (2011)	& 3080	\\ \hline
Battle of the Year (2013)	& 4074	\\ \hline
Cirque du Soleil: Journey of Man (2000)	& 897	\\ \hline
Creature from the Black Lagoon (1954)	& 680	\\ \hline
Dark Country (2009)	& 324	\\ \hline
Drive Angry (2011)	& 2437	\\ \hline
Exodus: Gods and Kings (2014)	& 5650	\\ \hline
Final Destination 5 (2011)	& 2212	\\ \hline
Flying Swords of Dragon Gate (2011)	& 2618	\\ \hline
Galapagos: The Enchanted Voyage (1999)	& 230	\\ \hline
Ghosts of the Abyss (2003)	& 946	\\ \hline
Hugo (2011)	& 3338	\\ \hline
Into the Deep (1994)	& 22	\\ \hline
Jack the Giant Slayer (2013)	& 5174	\\ \hline
Journey 2: The Mysterious Island (2012)	& 3184	\\ \hline
Journey to the Center of the Earth (2008)	& 1416	\\ \hline
Life of Pi (2012)	& 5160	\\ \hline
My Bloody Valentine (2009)	& 1627	\\ \hline
Oz the Great and Powerful (2013)	& 4559	\\ \hline
Pina (2011)	& 1827	\\ \hline
Piranha 3DD (2012)	& 1766	\\ \hline
\begin{tabular}[c]{@{}l@{}}Pirates of the Caribbean: \\ On Stranger Tides\end{tabular} (2011)	& 5015	\\ \hline
Pompeii (2014)	& 3644	\\ \hline
Prometheus (2012)	& 4188	\\ \hline
Sanctum (2011)	& 1976	\\ \hline
Saw 3D: The Final Chapter (2010)	& 2757	\\ \hline

\hline
\end{tabular}
    \caption{Stereo movies used in our experiments, part 1}
    \label{tab:3dmv-list1}
\end{table}

\begin{table}[h!]
    \centering
\begin{tabular}{|l|r|}
\hline
Name                                         & Frames \\ \hline\hline     

Sea Rex 3D (2010)	& 1015	\\ \hline
Silent Hill: Revelation 3D (2012)	& 1747	\\ \hline
Sin City: A Dame to Kill For (2014)	& 3585	\\ \hline
Space Station 3D (2002)	& 362	\\ \hline
Stalingrad (2013)	& 6453	\\ \hline
Step Up 3D (2010)	& 3209	\\ \hline
Step Up Revolution (2012)	& 3542	\\ \hline
Texas Chainsaw 3D (2013)	& 3089	\\ \hline
The Amazing Spider-Man (2012)	& 5378	\\ \hline
The Child's Eye (2010)	& 1232	\\ \hline
The Darkest Hour (2011)	& 3640	\\ \hline
The Final Destination (2009)	& 1998	\\ \hline
The Great Gatsby (2013)	& 4788	\\ \hline
\begin{tabular}[c]{@{}l@{}}The Hobbit:\\ An Unexpected Journey (2012)\end{tabular}	& 4128	\\ \hline
\begin{tabular}[c]{@{}l@{}}The Hobbit:\\ The Desolation of Smaug (2013)\end{tabular}	& 7266	\\ \hline
\begin{tabular}[c]{@{}l@{}}The Hobbit:\\ The Battle of the Five Armies (2014)\end{tabular}	& 6568	\\ \hline
The Hole (2010)	& 1685	\\ \hline
The Martian (2015)	& 4893	\\ \hline
The Three Musketeers (2011)	& 5284	\\ \hline
The Ultimate Wave Tahiti (2010)	& 638	\\ \hline
Ultimate G's (2000)	& 366	\\ \hline
Underworld: Awakening (2012)	& 3093	\\ \hline
X-Men: Days of Future Past (2014)	& 3482	\\ \hline
\hline
\textbf{Overall}                                                 & \textbf{146242}\\
\hline
\end{tabular}
    \caption{Stereo movies used in our experiments, part 2}
    \label{tab:3dmv-list2}
\end{table}

\section{Data}
\label{apdx:stereomovies_data}

\subsection{MiDaS Dataset Mixture}

\textbf{Stereo Movies} The original StereoMovies dataset collected in MiDaS~\cite{midas} consists of 23 stereo videos and features video frames from various non-static environments. We follow the similar data acquisition and processing protocol, but, we use 26 additional stereo movies, totalling 49 movies (listed in Tab.\ref{tab:3dmv-list1} and Tab.~\ref{tab:3dmv-list2}). The obtained video frames are highly diverse, containing landscapes, architecture, humans in action, and other various scenes.

We sample one frame per second from the collected videos. We omit the first and the last 10\% of frames as they usually contain opening and closing credits. We consider valid only the pixels where the discrepancy between left to right and right to left disparities is less than 8 pixels. Accordingly, we use only images where the disparity is valid for more than 80\% of pixels and the difference between maximal and minimal disparities exceeds 8 pixels.

\textbf{WSVD.} We do not use the WSVD dataset~\cite{wang2019web} used in the original MiDaS mixture since it contains data in the form of web links referring to the sources that have already been partially deleted.

\subsection{LeReS Dataset Mixture}

\textbf{DIML Outdoor.} DIML Outdoor contains calibrated and rectified stereo images so that disparities can be extracted via stereo matching. LeReS~\cite{LeReS} uses GANet~\cite{Zhang2019GANet}, while we perform stereo matching with AANet~\cite{xu2020aanet}.

\textbf{Holopix.} To obtain disparities from stereo data in Holopix~\cite{Holopix}, we opt for a more accurate PWCNet~\cite{pwcnet} instead of FlowNet2~\cite{IMKDB17} used in LeReS.

The other datasets in MiDaS and LeReS dataset mixtures are publicly available for downloading and contain ground truth data, so we use these datasets 'as is'.

\subsection{Data Used in Ablation Studies.} In ablation studies, we use the NYUv2 raw \cite{nyuv2} dataset. We select approximately 150k images from the training subset, evaluating on the original test subset of 654 images.

\begin{figure*}[ht!]
\centering
\setlength{\tabcolsep}{2pt}
\begin{tabular}{ccccc}
    \multirow{4}{*}[-30pt]{\shortstack{RGB\\\includegraphics[height=60pt]{images/supp/fig2/1113_source.png}}} & Mannequin~\cite{mannequin} & MegaDepth~\cite{megadepth}  & MiDaS~\cite{midas} &  \multirow{4}{*}[-30pt]{\shortstack{Ours, B5-LRN\\\includegraphics[height=60pt]{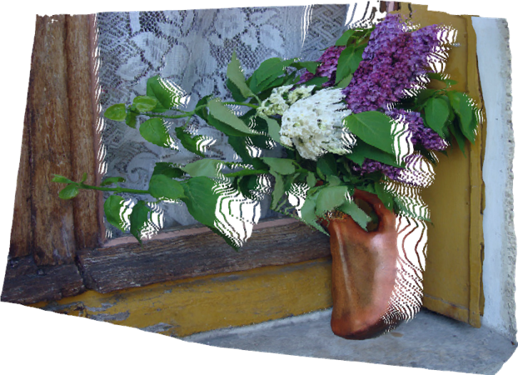}}} \\
    &
    \includegraphics[height=60pt]{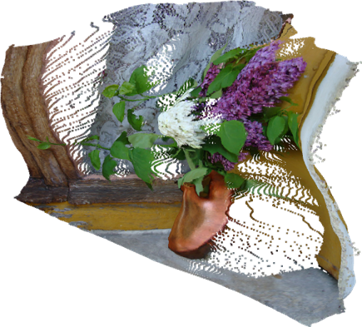} &
    \includegraphics[height=60pt]{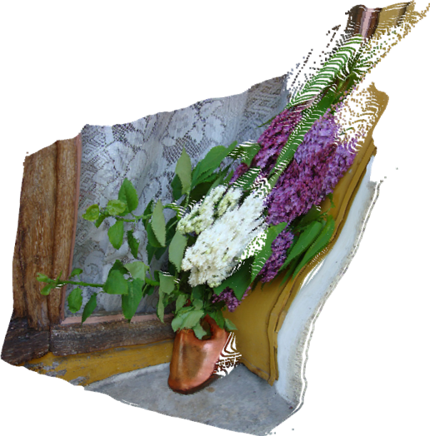} &
    \includegraphics[height=60pt]{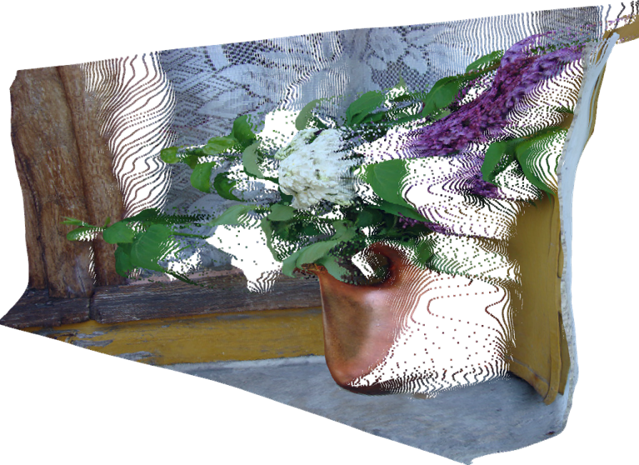} \\
    & DPT-Hybrid~\cite{DPT} & DPT-Hybrid-NYU~\cite{DPT} & LeReS+PCM~\cite{LeReS} \\
    &
    \includegraphics[height=60pt]{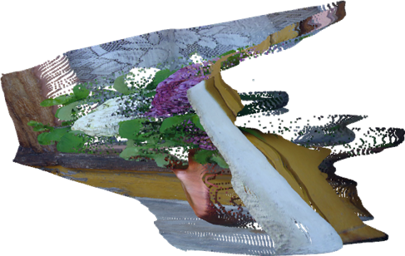} &
    \includegraphics[height=60pt]{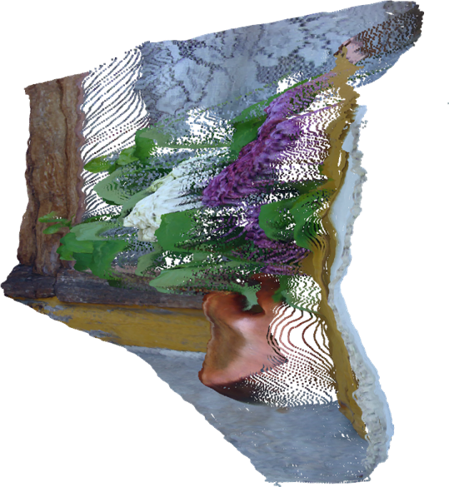} &
    \includegraphics[height=60pt]{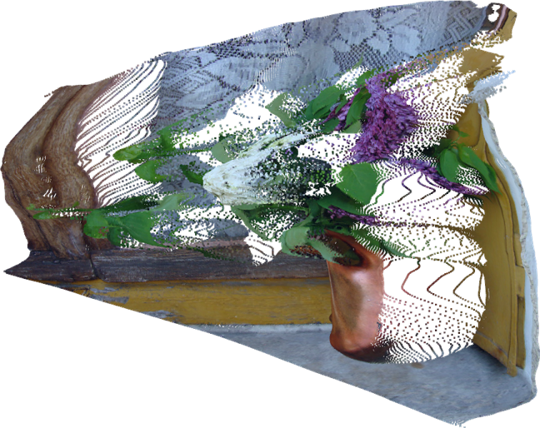} \\
    \\
    \\
    \multirow{4}{*}[-40pt]{\shortstack{RGB\\\includegraphics[height=60pt]{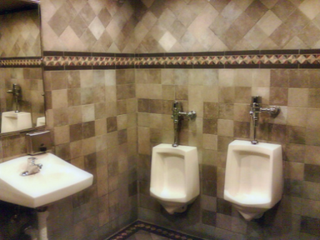}}} & Mannequin~\cite{mannequin} & MegaDepth~\cite{megadepth} &  MiDaS~\cite{midas} &\multirow{4}{*}[-40pt]{\shortstack{Ours, B5-LRN\\\includegraphics[height=60pt]{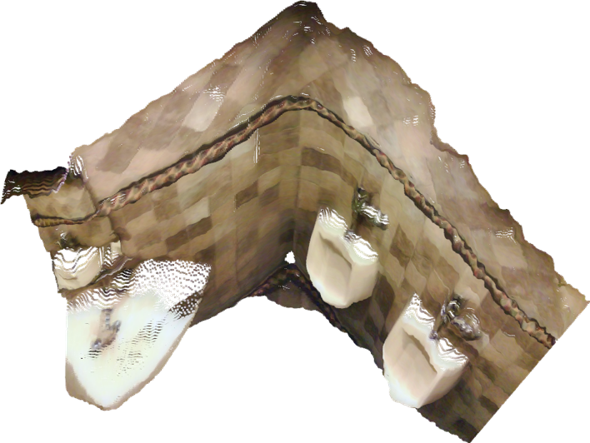}}} \\
     & \includegraphics[height=60pt]{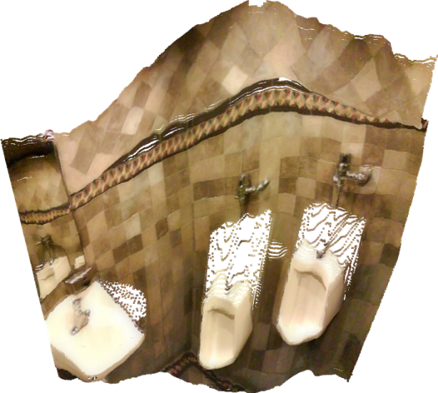} &
    \includegraphics[height=60pt]{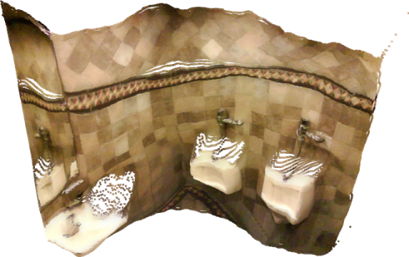} &
    \includegraphics[height=60pt]{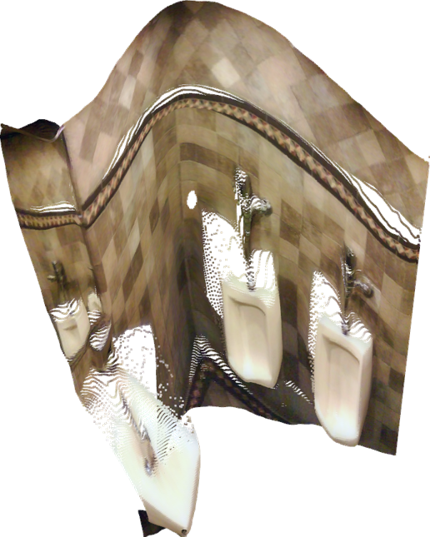} \\
    & DPT-Hybrid~\cite{DPT} & DPT-Hybrid-NYU~\cite{DPT} & LeReS+PCM~\cite{LeReS} \\
    &
    \includegraphics[height=60pt]{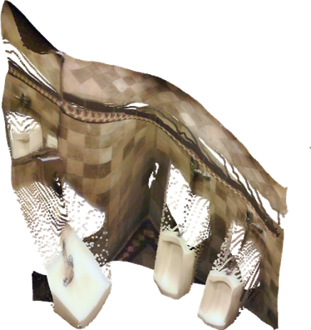} &
    \includegraphics[height=60pt]{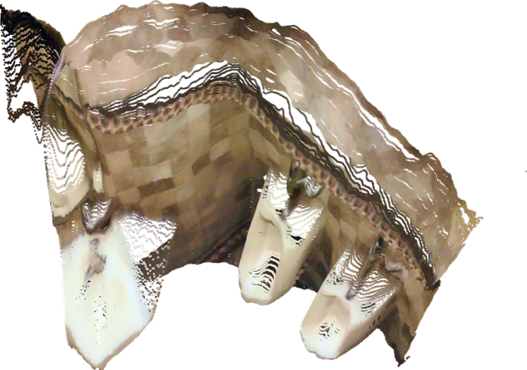} &
    \includegraphics[height=60pt]{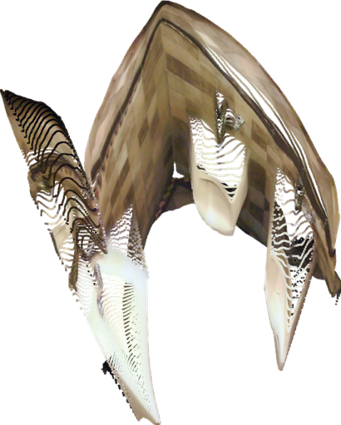} &\\
\end{tabular}
\caption{Point clouds reconstructed from depth estimates obtained with existing SVDE methods, including our GP\textsuperscript{2}-trained B5-LRN SVDE model.}
\label{fig:supp_point_clouds_all}
\end{figure*}

\begin{figure*}[h!]
\centering
\setlength{\tabcolsep}{2pt}
\begin{tabular}{cccc|cc}
    \multirow[c]{2}{*}{RGB} & \multirow[c]{2}{*}{MegaDepth~\cite{megadepth}} & \multirow[c]{2}{*}{MiDaS~\cite{midas}} & \multirow[c]{2}{*}{LeReS+PCM~\cite{LeReS}} &  B5-LRN & B5-LRN \\ 
    & & & & UTS only & GP\textsuperscript{2}-trained \\ 
    \includegraphics[height=75pt]{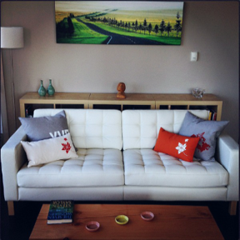} &
    \includegraphics[height=75pt]{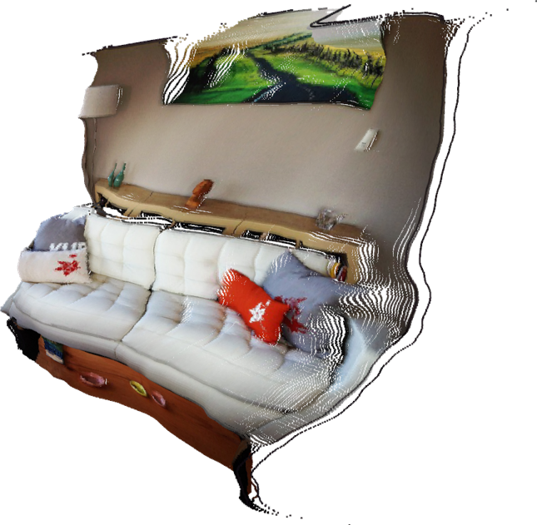} &
    \includegraphics[height=75pt]{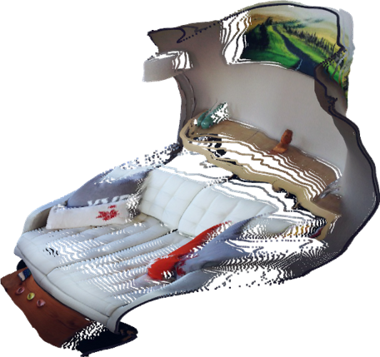} &
    \includegraphics[height=75pt]{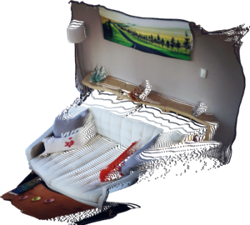} &
    \includegraphics[height=75pt]{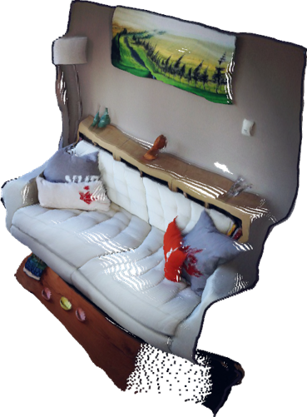} &
    \includegraphics[height=65pt]{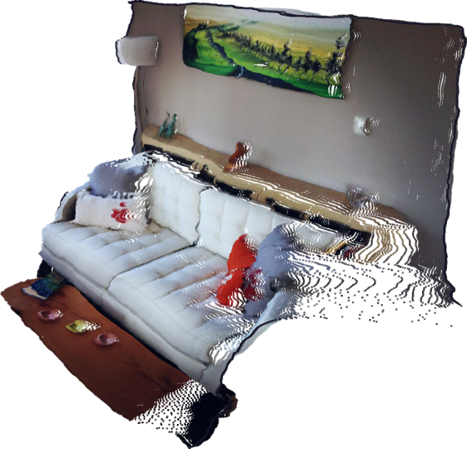} \\
    \includegraphics[height=65pt]{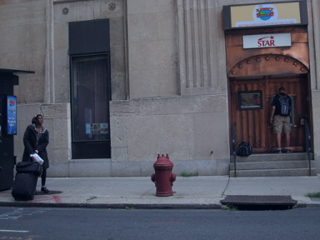} &
    \includegraphics[height=65pt]{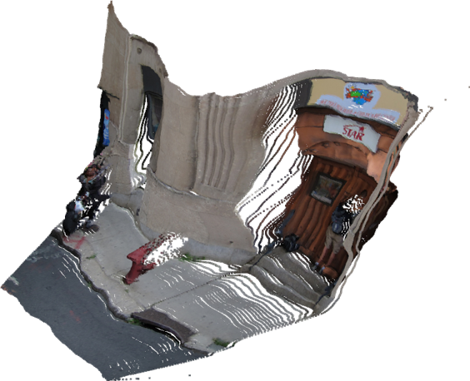} &
    \includegraphics[height=65pt]{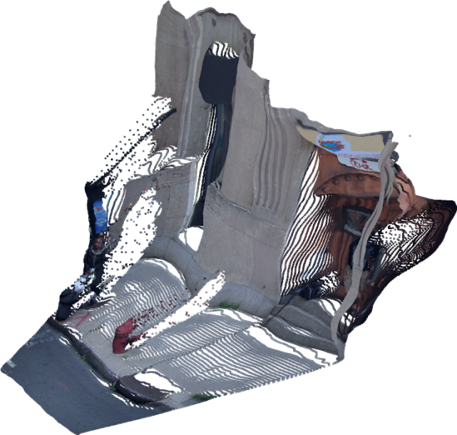} &
    \includegraphics[height=65pt]{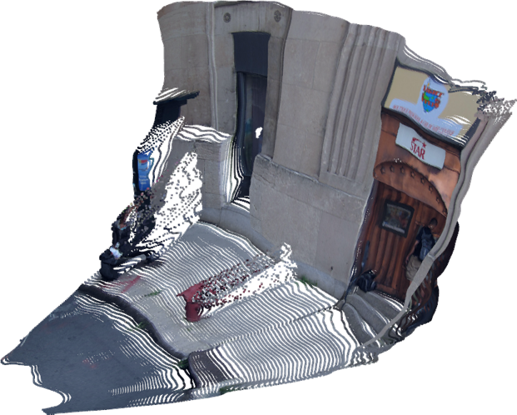} &
    \includegraphics[height=65pt]{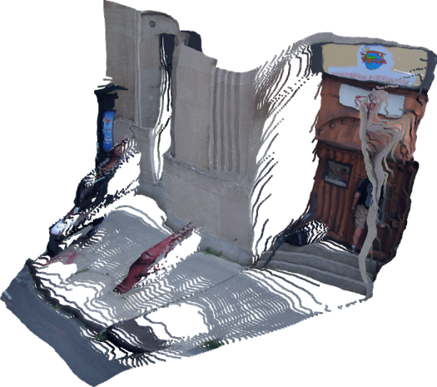} &
    \includegraphics[height=65pt]{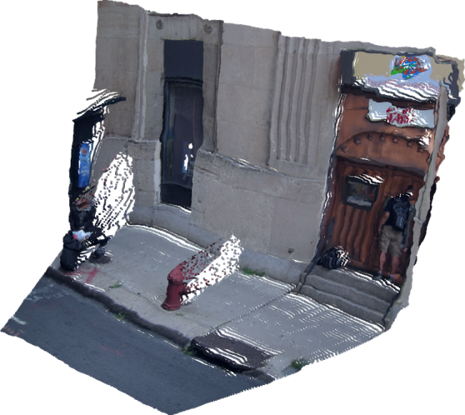} \\
\end{tabular}
\caption{Point clouds reconstructed from depth estimates obtained with existing SVDE methods, including our UTS-trained and GP\textsuperscript{2}-trained B5-LRN SVDE model.}
\label{fig:supp_point_clouds_uts}
\end{figure*}

\section{Visualizations}

In this section, we provide additional visualizations of point clouds reconstructed from depth maps estimated with different SVDE methods. To create a point cloud from predictions of our SVDE models, we apply the pre-trained focal recovery module from LeReS~\cite{LeReS}.

Qualitative comparison of our best GP\textsuperscript{2}-trained B5-LRN model with other existing SVDE methods is presented in Fig.~\ref{fig:supp_point_clouds_all}. Mannequin~\cite{mannequin} is trained on UTS data obtained from 3D reconstruction, so it predicts UTS depth. MiDaS~\cite{midas} predicts depth with incorrect shifts, resulting in severely distorted point clouds. 
DPT-Hybrid-NYU, which is fine-tuned on the training subset of NYUv2, provides better reconstructions than DPT-Hybrid trained only on the MIX6 dataset mixture~\cite{DPT}; nevertheless, both these methods fail to restore the actual scene geometry.

\begin{figure*}[!htbp]
\centering
\setlength{\tabcolsep}{2pt}
\begin{tabular}{ccc}
    RGB & LeReS+PCM~\cite{LeReS} & GP\textsuperscript{2}-trained B5-LRN \\
    \includegraphics[height=100pt]{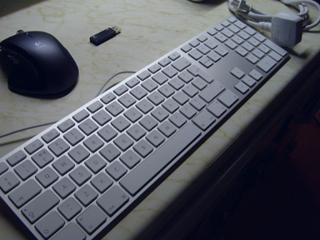} &
    \includegraphics[height=100pt]{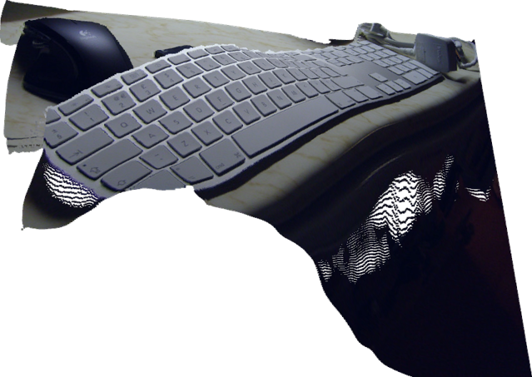} &
    \includegraphics[height=100pt]{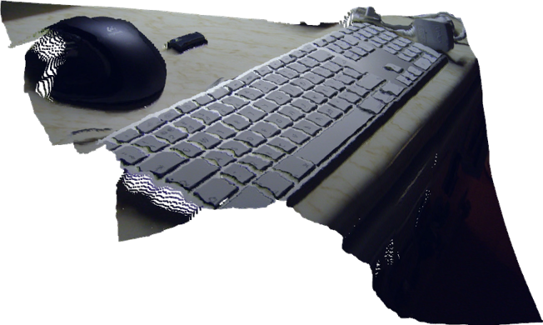} \\
    \includegraphics[height=100pt]{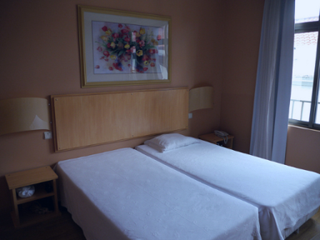} &
    \includegraphics[height=100pt]{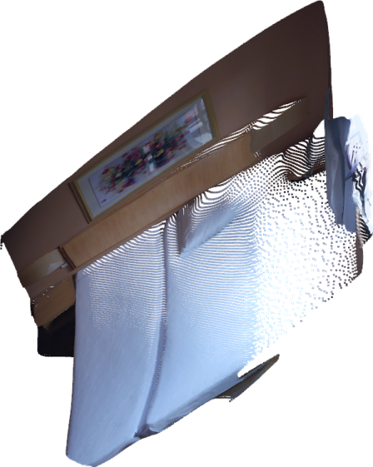} &
    \includegraphics[height=100pt]{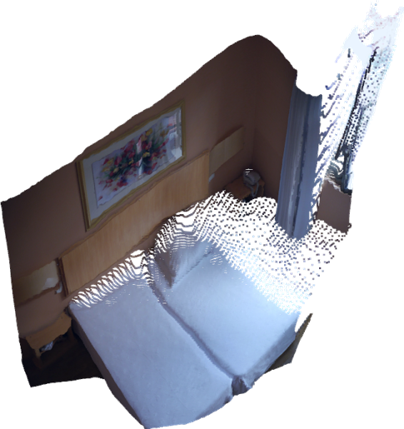} \\
    \includegraphics[height=100pt]{images/supp/fig3/1407_source.png} &
    \includegraphics[height=100pt]{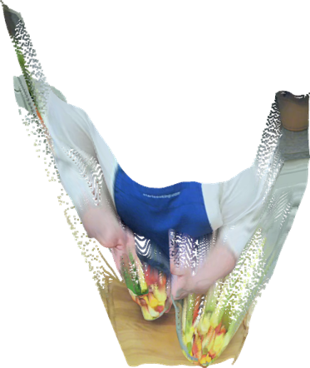} &
    \includegraphics[height=100pt]{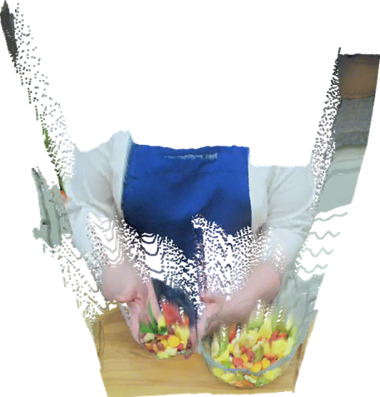} \\
    \includegraphics[height=100pt]{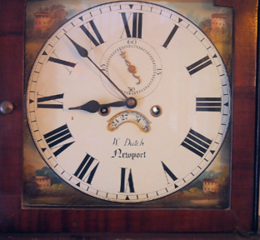} &
    \includegraphics[height=100pt]{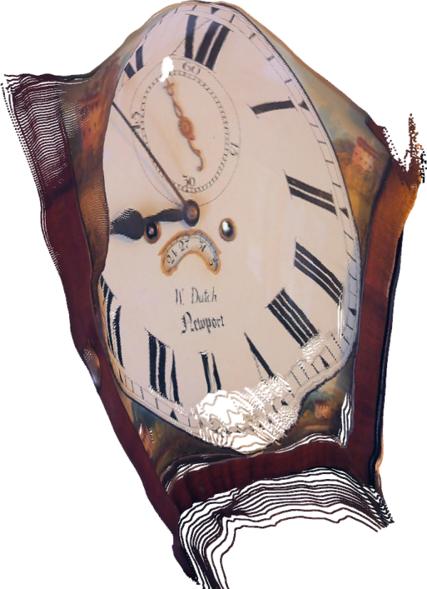} &
    \includegraphics[height=100pt]{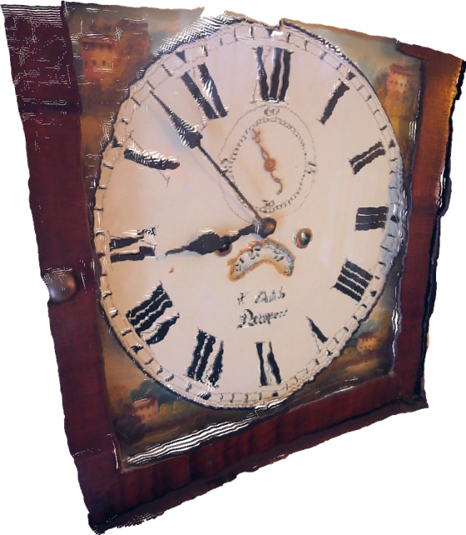} \\
    \includegraphics[height=100pt]{images/supp/fig3/1146_source.png} &
    \includegraphics[height=100pt]{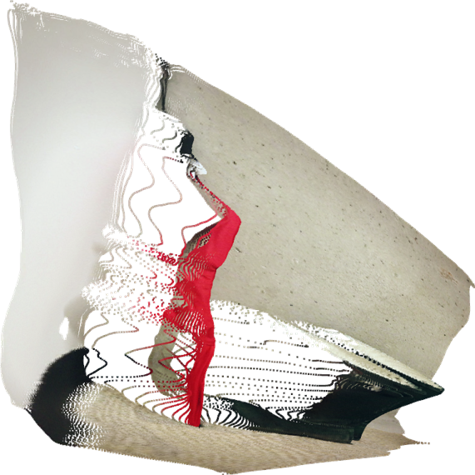} &
    \includegraphics[height=100pt]{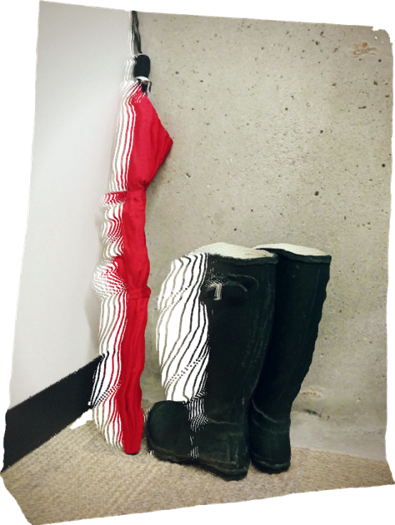} \\
    \includegraphics[height=100pt]{images/supp/fig3/1158_source.png} &
    \includegraphics[height=100pt]{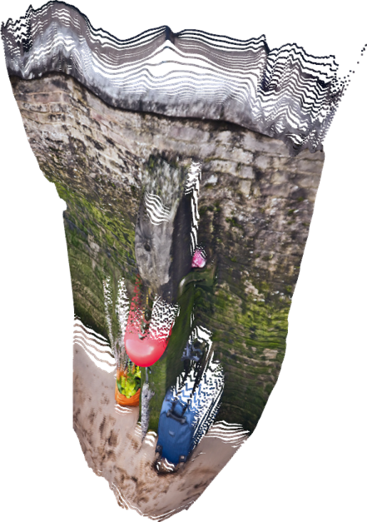} &
    \includegraphics[height=100pt]{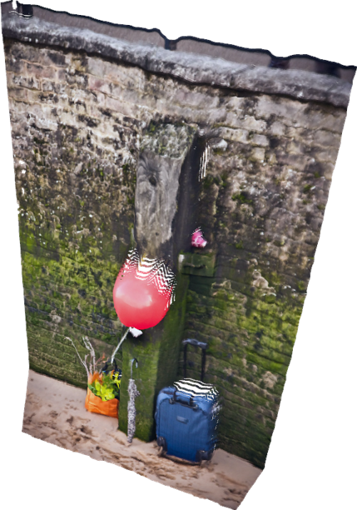} \\
\end{tabular}
\caption{Point clouds reconstructed from depth estimates obtained with LeReS+PCM and GP\textsuperscript{2}-trained B5-LRN.}
\label{fig:supp_point_clouds_leres}
\end{figure*}

\begin{figure*}[!htbp]
    \centering
    \includegraphics[width=\textwidth]{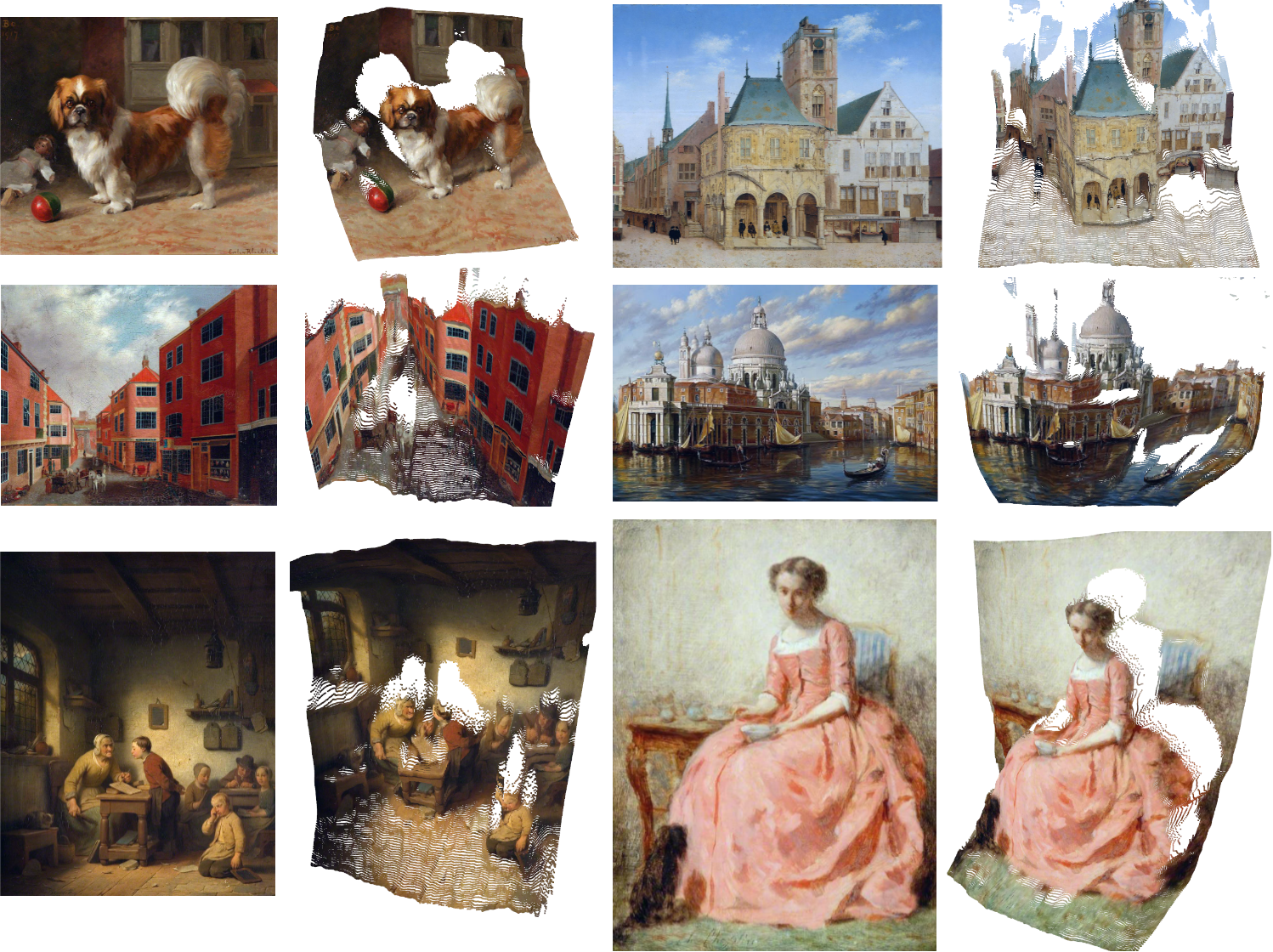}
    \caption{Point clouds obtained from depth estimates of our B5-LRN model. Paintings are a new data domain unseen during training, however, our method successfully handles these images, estimating depth adequately.}
    \label{fig:supp_art}
\end{figure*}

\begin{figure*}[!htbp]
    \centering
    \includegraphics[width=\textwidth]{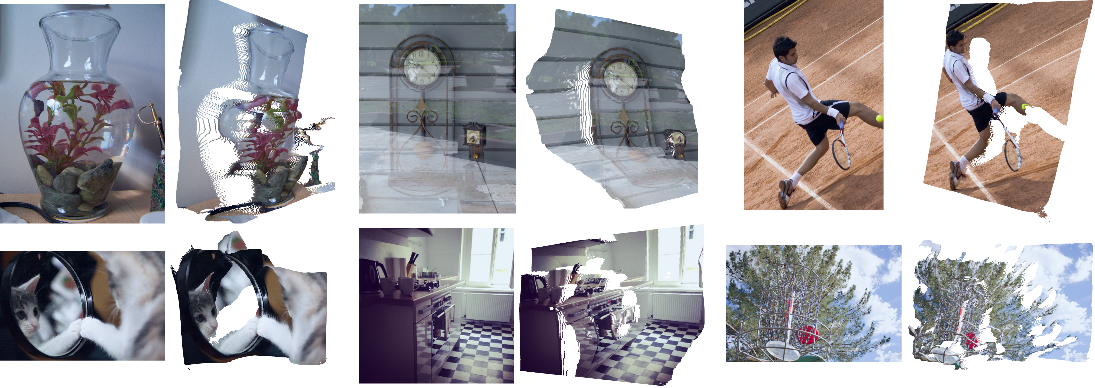}
    \caption{Failure cases of our B5-LRN model: as one might see, reflective and glassy surfaces, mirrors, thin objects are difficult for our model.}
    \label{fig:supp_failures}
\end{figure*}

In Fig.~\ref{fig:supp_point_clouds_uts}, we show the benefits of using UTSS data for training a geometry-preserving SVDE model. For this purpose, we train our B5-LRN model either on UTS data only or on a mixture of UTS and UTSS data using GP\textsuperscript{2}. According to the visualized point clouds, adding UTSS data to the training mixture improves the quality of reconstructions. We also compare our GP\textsuperscript{2}-trained B5-LRN model with geometry-preserving MegaDepth trained on only UTS data, non-geometry preserving MiDaS trained on a mixture of UTS and UTSS data, and geometry-preserving LeReS trained on a mixture of UTS and UTSS data. As one might see, our GP\textsuperscript{2}-trained B5-LRN allows recovering more accurate point clouds, which testifies in favor of the proposed training scheme.

To demonstrate the effectiveness of our training scheme for general-purpose geometry-preserving SVDE, we compare directly to LeReS+PCM being the only existing general-purpose and geometry-preserving SVDE method trained on a mixture of UTS and UTSS data. In Fig.~\ref{fig:supp_point_clouds_leres}, we visualize point clouds reconstructed using depth maps predicted by our GP\textsuperscript{2}-trained B5-LRN SVDE model and LeReS+PCM. Since both our model and LeReS use the same module for focal length estimation, the observed quality gap should be attributed to the better depth estimates obtained with our model.

Fig.~\ref{fig:supp_art} depicts the point clouds reconstructed from paintings by the GP\textsuperscript{2}-trained B5-LRN model. Paintings are a new data domain unseen during training; however, our model generalizes well even on non-photorealistic images.

Finally, we visualize some failures of GP\textsuperscript{2}-trained B5-LRN in Fig.~\ref{fig:supp_failures} to give a complete picture.
Expectedly, our model fails on highly reflective and glassy surfaces, mirrors, and thin objects.%

\end{document}